\documentclass[letterpaper, 10 pt, conference]{ieeeconf}  

\IEEEoverridecommandlockouts                              

\usepackage[bookmarks=true]{hyperref}
\usepackage{amsmath} 
\usepackage{multicol}
\usepackage{multirow}
\usepackage{todonotes}
\usepackage{amssymb}
\usepackage{array}
\usepackage{empheq}
\usepackage[caption=false,font=footnotesize]{subfig}
\usepackage{color}
\usepackage[export]{adjustbox}
\usepackage[noadjust]{cite}
\usepackage{graphicx} 

\usepackage{comment}

\newcommand{\tosay}{true}
\newcommand{\ludo}[1]{
{\ifthenelse{\boolean{\tosay}}{\begin{quotation}\textcolor{blue}{Ludo: #1}\end{quotation}}{}}}

\makeatletter
\newcommand{\thickhline}{%
    \noalign {\ifnum 0=`}\fi \hrule height 1.5pt
    \futurelet \reserved@a \@xhline
}
\newcolumntype{"}{@{\hskip\tabcolsep\vrule width 1pt\hskip\tabcolsep}}
\makeatother

\title{\LARGE \bf
    Robust Humanoid Contact Planning with Learned Zero- and One-Step Capturability Prediction
}

\author{
    Yu-Chi Lin$^{1}$, Ludovic Righetti$^{2,3}$, and Dmitry Berenson$^{1}$
    \thanks{$^{1}$University of Michigan, Ann Arbor, MI, USA, $^{2}$Max Planck Institute for Intelligent Systems, Tuebingen, Germany, $^{3}$New York University, New York, NY, USA. This work was supported in part by the Office of Naval Research under grant N000141712050}
}

\renewcommand{\baselinestretch}{.97}

\begin{document}

\maketitle
\thispagestyle{empty}
\pagestyle{empty}

\begin{abstract}

Humanoid robots maintain balance and navigate by controlling the contact wrenches applied to the environment. While it is possible to plan dynamically-feasible motion that applies appropriate wrenches using existing methods, a humanoid may also be affected by external disturbances. Existing systems typically rely on controllers to reactively recover from disturbances. However, such controllers may fail when the robot cannot reach contacts capable of rejecting a given disturbance. In this paper, we propose a search-based footstep planner which aims to maximize the probability of the robot successfully reaching the goal without falling as a result of a disturbance. The planner considers not only the poses of the planned contact sequence, but also alternative contacts near the planned contact sequence that can be used to recover from external disturbances. Although this additional consideration significantly increases the computation load, we train neural networks to efficiently predict multi-contact zero-step and one-step capturability, which allows the planner to generate robust contact sequences efficiently. Our results show that our approach generates footstep sequences that are more robust to external disturbances than a conventional footstep planner in four challenging scenarios. 
\end{abstract}

\section{Introduction}

\begin{figure}[t]
	\centering
	\includegraphics[scale=0.47]{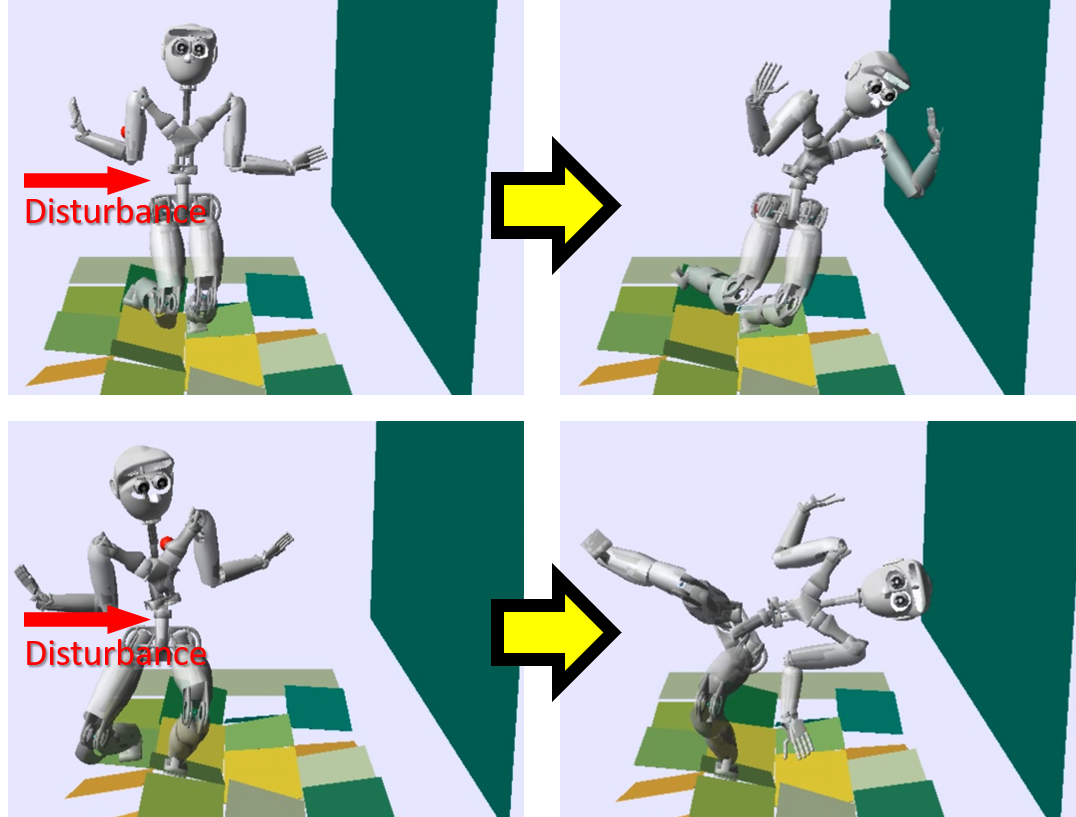}
	\caption{The robot walks over a rubble, and is impacted by a disturbance. Top: The robot walks close to the wall, and capture itself using a palm contact on the wall. Bottom: The robot cannot reach the wall, and falls down under the disturbance.}
	\label{intro_fig}
\end{figure}

Algorithms efficiently computing contact sequences
to traverse complex terrains are a fundamental building block for multi-contact behaviors of legged robots, in particular humanoids.
In order to reduce computational complexity, most contact planners generate contact sequences considering solely quasi-static constraints \cite{reachability_planner_journal,hauser,escande,CES,motion_mode_and_library}. However, a static stability criterion significantly decreases the set of possible contact transitions, which quickly leads to planning failure when attempting to traverse complex environments.
More recently, efficient planners using more general dynamic feasibility constraints have also been proposed \cite{croc,yuchi_icra_2019,centroidal_dynopt_1}.
Nevertheless, all these approaches assume fixed, deterministic environments and do not consider the robustness of contact sequences to potential environmental disturbances. In Figure \ref{intro_fig}, we show an example where the robot walks over rubble. There is a wall in the environment, and the robot can use palm contacts to capture itself against potential disturbances. However, without considering this information in the planner, a conventional contact planner could take the shortest feasible path which does not have access to the wall, and may cause the robot to fall down when a disturbance occurs.


In this paper, we propose a computationally efficient footstep planner that explicitly takes into account disturbances to increase motion robustness. In particular, we consider zero-step and one-step capture motions using either foot or palm contacts. 
Testing the existence of capture motions in multi-contact scenarios necessitates the solution to
a kino-dynamic optimal control problem \cite{zero_step_capturability_multi_contact,wang_contact_transition_tree}. However, it is prohibitively long to directly solve such problem in a footstep planner, as every candidate contact transition requires such a test.
Instead, we propose to train neural networks to predict the existence of a dynamically feasible capture motion using data generated offline with a kino-dynamic optimizer. The networks predict both zero-step and one-step capturability for a full-body dynamic model using both foot and palm capture motions.
We then query these networks in the footstep planning loop to inform the Anytime Non-parametric A*(ANA*) planner \cite{ana_star} about which footstep transitions are most robust to disturbances by measuring how many sampled contact poses can reject the disturbances. To the best of our knowledge, this work is the first footstep planner to use a learned model that predicts robot capturability under disturbances to produce more robust footstep sequences.

Our experiments first show that our neural networks achieve high accuracy in predicting robot capturability. We then compare our planning approach to a conventional distance-based footstep planner. Our results show that our approach generates footstep sequences that are more robust to external disturbances than the conventional method in four challenging scenarios. 

\section{Related Work}

Humanoid footstep planning has been broadly studied \cite{footstep1,footstep2,footstep3,footstep4,footstep5,footstep6,footstep7}. Conventional approaches plan a footstep sequence on a flat or piecewise-flat ground to avoid obstacles. To increase planning efficiency, these planners define a contact transition model which assumes all motions are dynamically feasible throughout the operating environment, and plan footstep sequences without explicitly checking their dynamic feasibility. Therefore, it is difficult for this kind of approach to generalize across environments.

More recent works verify dynamic feasibility by approximating it with quasi-static balance \cite{escande,hauser,reachability_planner_journal,motion_plan_library,CES}. These approaches are able to consider more diverse actions, including multi-contact motion. However, the quasi-static balance criterion is too conservative to consider dynamic motions.

To address the over-conservative nature of the quasi-static balance criterion, there are works which combine contact planning with dynamics optimization by solving a mixed integer convex program \cite{centroidal_dynopt_1,MICP_quadruped_1,MICP_quadruped_2}. This approach produces the global optimal solution, but does not generalize well to the complexity of the scene and the path length. To increase efficiency in checking dynamic feasibility, \cite{croc} proposes an approach by conservatively reformulating the problem as a linear program. Our previous work \cite{yuchi_icra_2019} proposes a graph-search based contact planner which uses neural networks to quickly predict the result of a dynamics optimizer. In this work, we further consider the robustness of the planned contact sequence under external disturbances.

Capturability analysis of the linear inverted pendulum (LIP) model was first proposed by \cite{capture_point}. Since then, it has been widely used to determine footstep placement in planning and control of robot dynamic walking \cite{Sugihara_LIP_capture,Takenaka_LIP_capture,Morisawa_LIP_capture,Englsberger_LIP_capture,Griffin_LIP_capture,khadiv2016step}. There are also works performing capturability analysis for the more complex variable-height inverted pendulum (VHIP) model to account for the height changes of the CoM. \cite{Pratt_VHIP_capture,Ramos_VHIP_capture,Koolen_VHIP_capture} address the balance control of humanoid robot using VHIP model for planar motions. \cite{caron2018capturability} further extends it to consider 3D movements, and develops an analytical tool to determine capturability in the VHIP model. \cite{zero_step_capturability_multi_contact} proposes an efficient analytical tool to compute zero-step capturability for multi-contact configuration using a centroidal dynamics model. However, it has strong assumptions on using zero angular momentum, and cannot generalize to use additional steps.

\section{Problem Statement}
We focus on the problem of planning humanoid footstep sequences considering the effect of external disturbances. Given an environment specified as a set of polygonal surfaces, an initial stance (set of poses of contacting end-effectors), a goal region, and a distribution of potential disturbances in the environment, we aim to output a dynamically feasible footstep sequence to move the robot from the initial stance to the goal region. In the planning, we consider not only where the robot can create contacts to achieve dynamically feasible motions, but also how well the robot can capture itself with existing and nearby contact locations, using \textbf{both feet and hands}, to reject disturbances sampled from the distribution of potential disturbances. \textcolor{black}{While it is important and desirable to generate those capture motions in real time, it is still an open problem and beyond the scope of this work.} Our goal is to find a footstep sequence that maximizes the probability of the robot reaching the goal successfully without falling as a result of a disturbance. \textcolor{black}{Notice that only feet are used in locomotion, but both feet and hands are available for rejecting potential disturbances.} We assume that the friction coefficient is given, as well as a fixed timing for each contact transition. In this work, we consider both zero-step and one-step capture motions.


	
	

\section{Iterative Kino-Dynamic Optimization}
\label{centroidal_momentum_opt}
In order to decide whether a capturing motion exists for the full robot model,
we use the kino-dynamic optimization method described in \cite{Herzog-2016b}.
\textcolor{black}{Given a sequence of collision-free contact poses, }the method decomposes the problem of optimizing dynamically-consistent whole-body motions and contact forces into 1) a dynamic optimization problem based on the centroidal dynamics \cite{OrinCentroidalMomentum} and 2) a kinematic optimization problem for the full-body motions. The algorithm computes the solution of each problems iteratively until both parts reach consensus over the center of mass $\mathbf{r}$, linear $\mathbf{l}$ and angular momentum $\mathbf{k}$ trajectories, leading to a locally optimal solution of the original problem.

In this work, we use the algorithm proposed in \cite{centroidal_dynopt_2} with fixed-time
to efficiently compute a solution for the dynamic optimization problem.
The centroidal dynamics expressed at the robot CoM is given by
\begin{equation}
\begin{bmatrix}
\mathbf{\dot{r}}\\ 
\mathbf{\dot{l}}\\ 
\mathbf{\dot{k}}
\end{bmatrix} = 
\begin{bmatrix}
\frac{1}{M}\mathbf{l}\\ 
M\mathbf{g} + \sum \mathbf{f}_{e}\\ 
\sum(T_{e}(\mathbf{z}_{e})-\mathbf{r})\times \mathbf{f}_{e} + \tau_{e}
\end{bmatrix}
\label{centroidal_dynamics}
\end{equation}
\noindent $M$ is the robot mass. $\mathbf{z}_{e}$ is the center of pressure (CoP) of each contact in the contact frame. $\mathbf{f}_{e}$ and $\tau_{e}$ are the contact force and torque at the CoP of each end-effector and finally, $T_{e}$ is a coordinate transform in the CoM frame. In addition to Eq.~\eqref{centroidal_dynamics}, contact forces need to be inside friction cones, and CoPs inside the support regions of each contact, to prevent the contact from sliding and tilting. 

To compute a dynamically robust motion we follow \cite{centroidal_dynopt_2} to minimize the weighted sum of the square norm of $\mathbf{l}$, $\mathbf{\dot{l}}$, $\mathbf{k}$, $\mathbf{\dot{k}}$, $\mathbf{f}_{e}$, and $\tau_{e}$. Lower values of $\mathbf{l}$ and $\mathbf{\dot{l}}$ help improve dynamic stability \cite{safe_locomotion}. Reducing $\mathbf{k}$ and $\mathbf{\dot{k}}$ help the robot perform more natural motion \cite{ang_mom_human_walking}. The $\mathbf{f}_{e}$ and $\tau_{e}$ terms encourage a more even distribution of forces and torques over all the contacts, which increases the controllability of the robot.
The dynamic optimizer is run before the kinematic optimizer. After the first iteration, torque limits are included in the dynamic optimizer by using the kinematic solution to find an approximation of the torque changes during the centroidal dynamics optimization. \textcolor{black}{To simplify the problem, in this work, collision avoidance is not considered in the optimization. In future works, we would like to incorporate the collision constraints using methods in \cite{trajopt}.}

A contact transition is considered capturable if the algorithm converges to consensus to a solution that satisfies all constraints after a maximum number of iterations, where we set constraints on the linear and angular momenta at the end of the movement to zero to ensure the robot will come to a stop.

\section{Modeling External Disturbances}

We model an external disturbance as an \textbf{instant} change in linear centroidal momentum. Therefore, an external disturbance $\mathbf{\delta}$ is a 3D vector: $\mathbf{\delta} \in \mathbb{R}^{3}$. We assume there is a known probability distribution of potential disturbances in each location $\mathbf{x} \in \mathbb{R}^{3}$ in the environment and the distribution is fixed during planning and execution time. To facilitate capturability checking, we discretize the distribution by sampling a set of representative disturbances from the distribution, and the probability of each disturbance sample is the total probability integrated over the Voronoi cell of the disturbance sample. Let $D(\mathbf{x})$ be the set of all representative disturbances. We assume that for any short period of time $T$, there will only be one disturbance, so we have
\begin{equation}
\sum_{i=1}^{N_{D}(\mathbf{x})} P\left(\mathbf{\delta}_{i},T\right) = 1, \ D(\mathbf{x}) = \left\{\mathbf{\delta}_{i}\left|i=1,2,\hdots,N_{D}(\mathbf{x})\right.\right\}
\label{disturbance_rel}
\end{equation}

\noindent where $P\left(\mathbf{\delta}_{i},T\right)$ is the probability that $\mathbf{\delta}_{i}$ happens once within time duration $T$, and $N_{D}(\mathbf{x})$ is the number of disturbance samples in $D(\mathbf{x})$.

\section{Evaluation of Capturability}
\label{evaluation_of_capturability}


To evaluate capturability, we adopt the approach of iterative kino-dynamic optimization described in Section \ref{centroidal_momentum_opt}. Since we model the disturbance as an instant change in linear momentum, we use the post-disturbance centroidal dynamics state $[\mathbf{r}_{0}, \mathbf{l}_{0}, \mathbf{k}_{0}]^{T}$, the centroidal dynamics state immediately after the disturbance $\mathbf{\delta}$, as the initial state of the iterative kino-dynamic optimization, and define it as $[\mathbf{r}_{0}, \mathbf{l}_{0}, \mathbf{k}_{0}]^{T} = [\mathbf{r}_{b}, \mathbf{l}_{b}+\delta, \mathbf{k}_{b}]^{T}$, where $[\mathbf{r}_{b}, \mathbf{l}_{b}, \mathbf{k}_{b}]^{T}$ is the centroidal dynamics state before disturbance.

In this work, we consider two kinds of capture motions: zero-step capture (capturing without making new contacts), and one-step capture (capturing by making one new contact). For zero-step capture, the initial condition of the optimization includes $[\mathbf{r}_{0}, \mathbf{l}_{0}, \mathbf{k}_{0}]^{T}$, and existing contact poses. For one-step capture, in addition to the above initial conditions, we also specify a target contact pose for one of the free end-effectors. 

To determine capturability, we first optimize the initial kinematic states $[\mathbf{q}_{0},\mathbf{\dot{q_{0}}}]$ to track $[\mathbf{r}_{0}, \mathbf{l}_{0}, \mathbf{k}_{0}]^{T}$ and the existing stance $S$ (set of contacting end-effectors poses), and then run kino-dynamic optimization for three iterations. If a kino-dynamically feasible solution can be found such that the linear and angular momentum converge to zero at the end of the motion, then the robot is capturable under the specified initial conditions: $(\mathbf{r}_{0}, \mathbf{l}_{0}, \mathbf{k}_{0}, S)$. For the one-step capture case, we try three different durations for the robot to move the end-effector to make contact: 0.2, 0.4 and 0.6 seconds. If any duration is feasible, then the robot is capturable given the initial conditions. \textcolor{black}{The evaluation for each contact pose takes from the order of 100 ms to 1 s depending on the difficulty of the situation and the number of iterations attempted. Although it is prohibitively long to be included in a planning loop, we can collect the result offline, and fit it with an computationally efficient model.}

\begin{figure}[tp]
	\centering
	\includegraphics[scale=0.45]{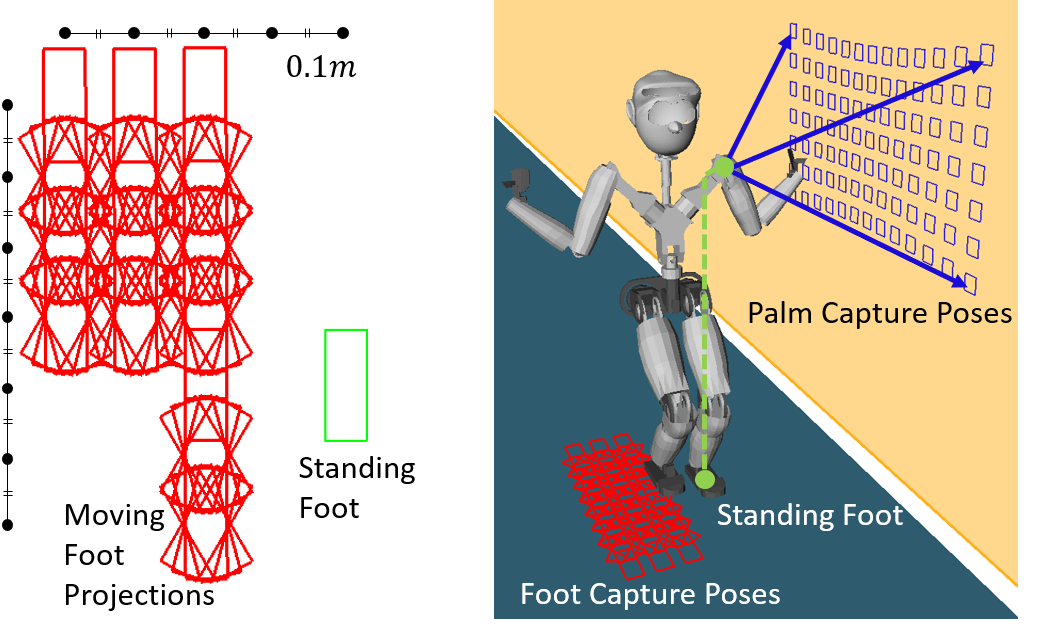}
	\vspace{-0.15in}
	\caption{(a) Left: Foot contact transition model in searching contact sequence, (b) Right: Possible foot and palm contact projections for one-step capture motion given the standing foot pose. \textcolor{black}{The projections are shown on flat surfaces as an illustrative example. When generating training data we sample contact poses with random tilt angles.}}
	\label{contact_projection}
	\vspace{-0.25in}
\end{figure}



\begin{figure}[t]
\subfloat{
    \adjustbox{width=0.51\columnwidth,valign=B}{
    \begin{tabular}{|c|c|l|c|}
\hline
Index & \begin{tabular}[c]{@{}c@{}}Capture \\ Motion Type\end{tabular}               & \multicolumn{1}{c|}{\begin{tabular}[c]{@{}c@{}}Capture \\ Motion\end{tabular}} & \begin{tabular}[c]{@{}c@{}}Input\\ Dim.\end{tabular} \\ \hline
0     & \begin{tabular}[c]{@{}c@{}}Zero-Step \\ Capture\end{tabular}                 & \begin{tabular}[c]{@{}l@{}}Maintain one \\ foot contact\end{tabular}           & 12                                                   \\ \hline
1     & \multirow{3}{*}{\begin{tabular}[c]{@{}c@{}}One-Step \\ Capture\end{tabular}} & \begin{tabular}[c]{@{}l@{}}Make the other \\ foot contact\end{tabular}         & 18                                                   \\ \cline{1-1} \cline{3-4} 
2     &                                                                              & \begin{tabular}[c]{@{}l@{}}Make the same \\ side palm contact\end{tabular}     & 18                                                   \\ \cline{1-1} \cline{3-4} 
3     &                                                                              & \begin{tabular}[c]{@{}l@{}}Make the opposite \\ side palm contact\end{tabular} & 18                                                   \\ \hline
\end{tabular}}
}
\subfloat{
    \includegraphics[scale=0.22]{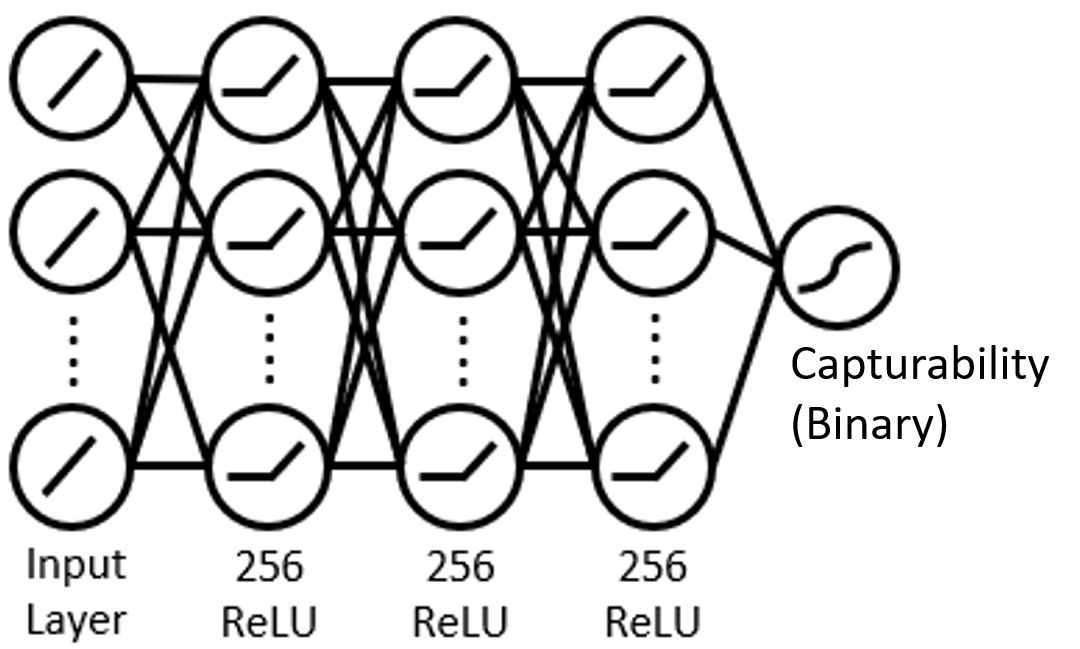}
}
\vspace{-0.05in}
\caption{Left: Capture motions considered in this work and their feature dimension. Every capture motion initially has one foot contact, and the side of the palm contacts is relative to the standing foot side. Right: The network structure to predict capturability. The learning rate is $5\times10^{-5}$ and there are dropout layers between fully-connected layers with $0.1$ dropout rate.}
\label{capture_motion_case_and_network}
\end{figure}

\section{Learning the Result of the Kino-Dynamic Optimization of Capture Motions}

For each contact transition evaluated in contact planning, the planner needs to decide if the robot can capture itself under a set of disturbances $D$, and for each disturbance $\mathbf{\delta}_{i} \in D$, many potential contacts may be considered to capture the robot in one step. Therefore, it is computationally prohibitive to run the iterative kino-dynamic optimization in the planning loop. To reduce online computation, we train a set of neural network classifiers offline to determine capturability. Each neural network corresponds to a separate capture motion involving different contacts, as shown in Figure \ref{capture_motion_case_and_network}.

The classifiers predict whether the optimizer can find a kino-dynamically feasible solution to capture the robot given the initial conditions described in Section \ref{evaluation_of_capturability}. Since angular momenta are generally low in walking motion \cite{ang_mom_human_walking}, we assume $\mathbf{k}_{0}=0$ and do not include it as the input of the network to improve data efficiency. \textcolor{black}{As shown in Figure \ref{capture_motion_case_and_network}, the classifiers take the initial standing foot pose, the capture contact pose, and $[\mathbf{r}_{0}, \mathbf{l}_{0}]^{T}$ as inputs, and have a 1D binary output, which represents whether the optimizer can find a kino-dynamically feasible solution to capture the robot.} Because most humanoid robots have symmetric kinematic structures, we utilize this symmetry and define 4 kinds of capture motion, as shown in Figure \ref{capture_motion_case_and_network}. For zero-step capture cases, the involved contact poses are only the existing contact poses; for one-step capture cases, a new contact pose of a free end-effector is considered. Each contact pose is a $\mathbb{R}^{6}$ vector which consists of position and orientation in Tait-Bryan angles, convention X-Y-Z, in $[-\pi, \pi)$. To capture the spatial relationship of the orientation with angles near $\pm\pi$, we duplicate those samples with $\mp2\pi$ in the training data.

\textcolor{black}{To collect meaningful training data, we determine the sampling space based on the robot's reachability and the target application. If we randomly sample contact poses in a wide space, most samples are not feasible or not useful for our application. To address this issue, we first get a rough estimate of the robot's reachability with kinematic optimizers on a set of widely-sampled contact poses, and then reduce the sampling space by defining sampling intervals in each dimension of $SE(3)$ to focus more on the robot's reachable space and poses required by the application. Although the training data will need to be recollected if different robots or applications are considered, in this way, we can get a more balanced data set.}

In this work, we collect data by sampling the initial standing foot contact pose with random tilt angles within $\pm25^{\circ}$ from Z axis. $\mathbf{r}_{0}$ is randomly sampled relative to the foot pose based on the robot's reachability, and $\mathbf{l}_{0}$ is randomly sampled in the magnitude interval of $m[0,1]\text{kg}\cdot\text{m/s}$, where $m$ is the robot mass, and its orientation is randomly sampled within $\pm45^{\circ}$ from the XY plane. For one-step capture cases, we sample capture contact poses using models shown in Figure \ref{contact_projection}. Each contact is projected with randomly selected depth and tilt angle to form a diverse set of initial conditions. Each sampled initial condition is supplied to the kino-dynamic optimizer described in Section \ref{evaluation_of_capturability} to decide its label. A different neural network is trained to determine capturability for each type of capture motion, but we use the same network structure for all capture motions to simplify the implementation, as shown in Figure \ref{capture_motion_case_and_network}.

\section{Anytime Discrete-Search Contact Planner}

We formulate the contact planning problem as a graph search problem. Each state $s$ in the graph is represented by a set of: a stance $S(s)$, a CoM position $\mathbf{r}(s)$, and a linear momentum $\mathbf{l}(s)$. Each action is a foot contact transition, which means moving one foot to a new pose. Contact transitions are predefined as a discrete set of foot projections, shown in Figure \ref{contact_projection}(a), and we adopt the contact projection approach in \cite{traversability_humanoids}. 

For each contact transition $\varepsilon(s,s')$ from state $s$ to state $s'$, the planner generates a new state with a stance which differs from the current stance by the moving contact pose. We assume there is a 0.4 second long swing phase followed by 0.6 second double support phase for each contact transition. We follow our previous work \cite{yuchi_icra_2019}, given $S(s)$, $S(s')$, $\mathbf{r}(s)$ and $\mathbf{l}(s)$, we use neural networks to predict dynamic feasibility of the contact transition, and determine $\mathbf{r}(s')$ and $\mathbf{l}(s')$.

We solve the contact planning problem with Anytime Non-parametric A*(ANA*) algorithm \cite{ana_star}. ANA* is an anytime variation of the A* algorithm. It initially inflates the heuristic and determines which node to expand mainly by evaluating its heuristic. Once a solution is found, it then reduces the inflation of the heuristic, and improves the solution over time. In this way, a feasible solution can be generated quickly, and helps reduce the search space to find a better solution over time. The cost of each action connecting two states $s$ and $s'$ is defined as
\begin{equation}
\Delta g(s,s') = d(s,s') + w_{s} + w_{cap}c_{cap}(s,s')
\label{capture_edge_cost}
\end{equation}

\noindent where $d(s,s')$ is the euclidean XY distance between the mean foot positions of state $s$ and $s'$, $w_{s}$ is a fixed step cost, and $c_{cap}$ is the capturability cost and $w_{cap}$ is its corresponding weight. We aim to generate a contact sequence which maximizes the robot's success rate to reach the goal without falling due to disturbance. Therefore, $c_{cap}$ should be determined by the probability that the robot can capture itself during the contact transition $\varepsilon(s,s')$ from $s$ to $s'$ given the probability distribution of the disturbances. We denote the capture probability as $P_{\text{success}}\left(\varepsilon(s,s')\right)$.

\begin{figure}[tp]
	\centering
	\includegraphics[scale=0.4]{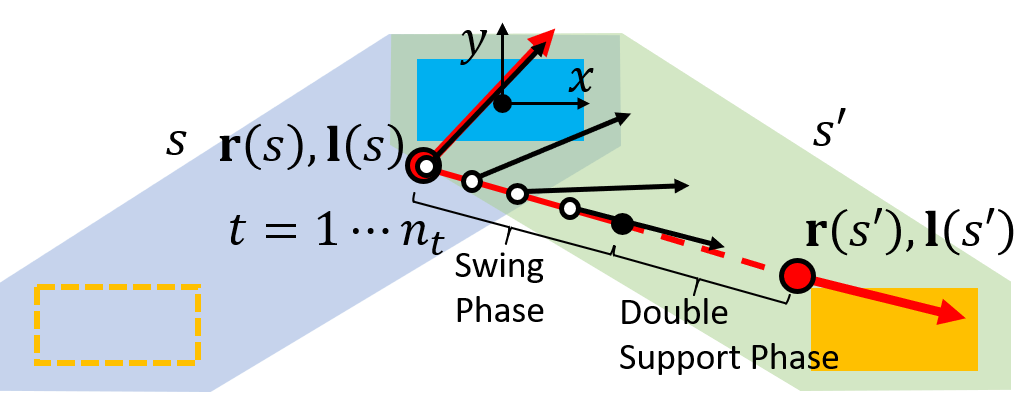}
	\caption{Approximated CoM position and linear momentum used to check capturability in Swing Phase Discretization. Blue and yellow boxes represent standing and swing foot, respectively. In practice, we let $n_{t}=4$ to represent 4 time steps in the swing phase: $0^{+},0.1,\cdots,0.3$ seconds from the start of the swing.}
	\label{com_traj}
\end{figure}

To determine $P_{\text{success}}\left(\varepsilon(s,s')\right)$, we consider two different approaches:
\begin{itemize}
  \item Swing Phase Discretization: Considering $n_{\mathbf{t}}$ pairs of $(\mathbf{r},\mathbf{l})$ from discretized time steps during the swing phase of contact transition from $s$ to $s'$, as shown in Figure \ref{com_traj}.
  \item Worst-case CoM Estimate: Considering only the $(\mathbf{r},\mathbf{l})$ pair right after the robot breaks a contact to start the swing phase (approximated as $(\mathbf{r}(s),\mathbf{l}(s))$).
\end{itemize}

For Swing Phase Discretization, $P_{\text{success}}\left(\varepsilon(s,s')\right)$ is defined as
\begin{equation}
\begin{split}
& \prod_{t=1}^{n_{t}} \sum_{i=1}^{N_{D}(\mathbf{r}_{t})}P_{\text{reject}}\left(\mathbf{r}_{t},\mathbf{l}_{t},S_{\text{swing},\varepsilon(s,s')},\delta_{i}\right)P\left(\mathbf{\delta}_{i},\frac{0.4}{n_{t}}\right) \\
& \left\{\begin{matrix}
\mathbf{r}_{t} = \frac{n_{t}-t}{n_{t}-1}\mathbf{r}(s) + \frac{t-1}{n_{t}-1}\mathbf{r}_{\text{swing},\varepsilon(s,s')} \\ 
\mathbf{l}_{t} = \frac{n_{t}-t}{n_{t}-1}\mathbf{l}(s) + \frac{t-1}{n_{t}-1}\mathbf{l}_{\text{swing},\varepsilon(s,s')} 
\end{matrix}\right. t=1,\cdots, n_{t}
\end{split}
\end{equation}

\noindent where $P_{\text{reject}}\left(\mathbf{r}_{t},\mathbf{l}_{t},S_{\text{swing},\varepsilon(s,s')},\delta_{i}\right)$ means the probability of the robot rejecting disturbance $\delta_{i} \in D(\mathbf{r}(s))$ with centroidal dynamics state before disturbance $[\mathbf{r}_{b}, \mathbf{l}_{b}, \mathbf{k}_{b}]^{T}=[\mathbf{r}_{t}, \mathbf{l}_{t}, 0]^{T}$, and the robot's stance in swing phase $S_{\text{swing},\varepsilon(s,s')}$. $\mathbf{r}_{\text{swing},\varepsilon(s,s')}$ and $\mathbf{l}_{\text{swing},\varepsilon(s,s')}$ are $\mathbf{r}$ and $\mathbf{l}$ at the end of the swing phase, and they are set empirically to be $\mathbf{r}_{\text{swing},\varepsilon(s,s')} = 0.4\mathbf{r}(s') + 0.6\mathbf{r}(s)$ and $\mathbf{l}_{\text{swing},\varepsilon(s,s')}=\mathbf{l}(s')$, respectively. Although only time steps in swing phase are considered here, empirically we find that for each step cycle, the robot has similar performance to reject disturbances by reactive stepping in double support phase or one-step capture in swing phase. Therefore, to reduce the computation load, we sample only from the swing phase in planning.

For Worst-case CoM Estimate, $P_{\text{success}}\left(\varepsilon(s,s')\right)$ is defined as
\begin{equation}
\sum_{i=1}^{N_{D}(\mathbf{r}(s))}P_{\text{reject}}\left(\mathbf{r}(s),\mathbf{l}(s),S_{\text{swing},\varepsilon(s,s')},\delta_{i}\right)P\left(\mathbf{\delta}_{i},0.4\right)
\end{equation}

\noindent In this definition, $P_{\text{success}}\left(\varepsilon(s,s')\right)$ only depends on $s$ and $S_{\text{swing},\varepsilon(s,s')}$, so for all $s'$ with the same $S_{\text{swing},\varepsilon(s,s')}$, $P_{\text{success}}\left(\varepsilon(s,s')\right)$ is the same. Therefore, compared to Swing Phase Discretization, Worst-case CoM Estimate reduces the computation time significantly because it only considers one centroidal dynamics state. \textcolor{black}{During the contact transition, disturbances pushing toward $+y$ direction in standing foot frame are hard to capture with the swing foot because of the kinematic constraints. As seen in Figure \ref{com_traj}, we observe that in dynamic walking, at the start of the swing phase, the robot has the highest $+y$ component of the linear momentum. Therefore, in Worst-case CoM Estimate, we sample the start of the swing phase of each $\varepsilon(s,s')$, and use it to determine $P_{\text{success}}\left(\varepsilon(s,s')\right)$.}

\subsection{Modelling disturbance rejection probability}
\label{dusturbance_rejection_probability}
Both definitions of $P_{\text{success}}\left(\varepsilon(s,s')\right)$ require the disturbance rejection probability $P_{\text{reject}}\left(\mathbf{r},\mathbf{l},S_{\text{swing},\varepsilon},\delta\right)$. For each $S_{\text{swing},\varepsilon}$, we use the foot and palm projection model shown in Figure \ref{contact_projection}(b) to find all possible capture poses. We then query the neural networks with $\mathbf{r}$, $\mathbf{l}$, $S_{\text{swing},\varepsilon}$ and each of those capture poses, and count the number of queries that output ``capturable'', including the zero-step capture motion, denoted as $n_{c}$. \textcolor{black}{Since the neural networks simplify the capturability check by abstracting the initial kinematics state to be a combination of a stance and a dynamics state, and assuming no initial angular momentum, we expect errors caused by these simplifications. Therefore, we would like to improve the planner robustness by favoring transitions $\varepsilon(s,s')$ which are predicted by the networks to be capturable with more capture poses (higher $n_{c}$ for each disturbance). Therefore, }we model $P_{\text{reject}}\left(\mathbf{r},\mathbf{l},S_{\text{swing},\varepsilon},\delta\right)$ as $1 - \exp(-\gamma n_{c})$, where $\gamma \in \mathbb{R}^{+}$ is a user defined constant. This model captures the idea that the robot is more likely to reject the disturbance if more network queries with different capture poses determine the condition to be capturable.


\subsection{Capturability Cost}

For a path $T_{cp}$ (a sequence of $K$ contact transitions), the probability that the robot finishes the path without falling due to external disturbance is
\begin{equation}
P_{\text{success}}\left(T_{cp}\right) = \prod_{k=1}^{K}P_{\text{success}}\left(\varepsilon_{k}\right)
\end{equation}
\noindent where $\varepsilon_{k}$ is the $k$th contact transition in $T_{cp}$. Our goal is to maximize $P_{\text{success}}\left(T_{cp}\right)$, which can be achieved by minimizing $\sum_{k=1}^{K}-\text{log}\left(P_{\text{success}}\left(\varepsilon_{k}\right)\right)$. Therefore, we define $c_{cap}$ as
\begin{equation}
c_{cap}(s,s') = -\text{log}\left(P_{\text{success}}\left(\varepsilon(s,s')\right)\right)
\end{equation}
\noindent With this definition of $c_{cap}$, we can find a path with maximum success rate by minimizing the total capturability cost of the path, which is done by the ANA* algorithm. In practice, we set $w_{cap} \gg w_{s}, d(s,s')$ to let ANA* focus on maximizing $P_{\text{success}}\left(T_{cp}\right)$.

\subsection{Contact Planning Heuristic}

To guide the search, we follow \cite{yuchi_icra_2019} and define the heuristic function by computing a policy for a simplified robot model moving on an SE(2) grid. The robot simplified model is a floating box. We first prune out every cell in the grid where there is no ground or there is collision between the box and the environment, and plan with Dijkstra's algorithm from the goal cell using an 8-connected grid transition model. By doing so, every cell connected to the goal cell will get a shortest distance $d_{\text{Dijkstra}}(s)$ to reach the goal and a policy which indicates the neighboring cell to go to. During contact planning, the planner queries this policy with the mean foot position on the XY plane, and the mean foot rotation about the Z axis to compute the heuristic.
\begin{equation}
h(s) = d_{\text{Dijkstra}}(s) + w_{s}\frac{d_{\text{Dijkstra}}(s)}{\Delta d_{max}}
\label{heuristics}
\end{equation}

\noindent where $\Delta d_{max}$ is an overestimate of the maximum distance the mean foot pose can travel in one transition.

\begin{figure*}[tp]
	\centering
	\includegraphics[scale=0.28]{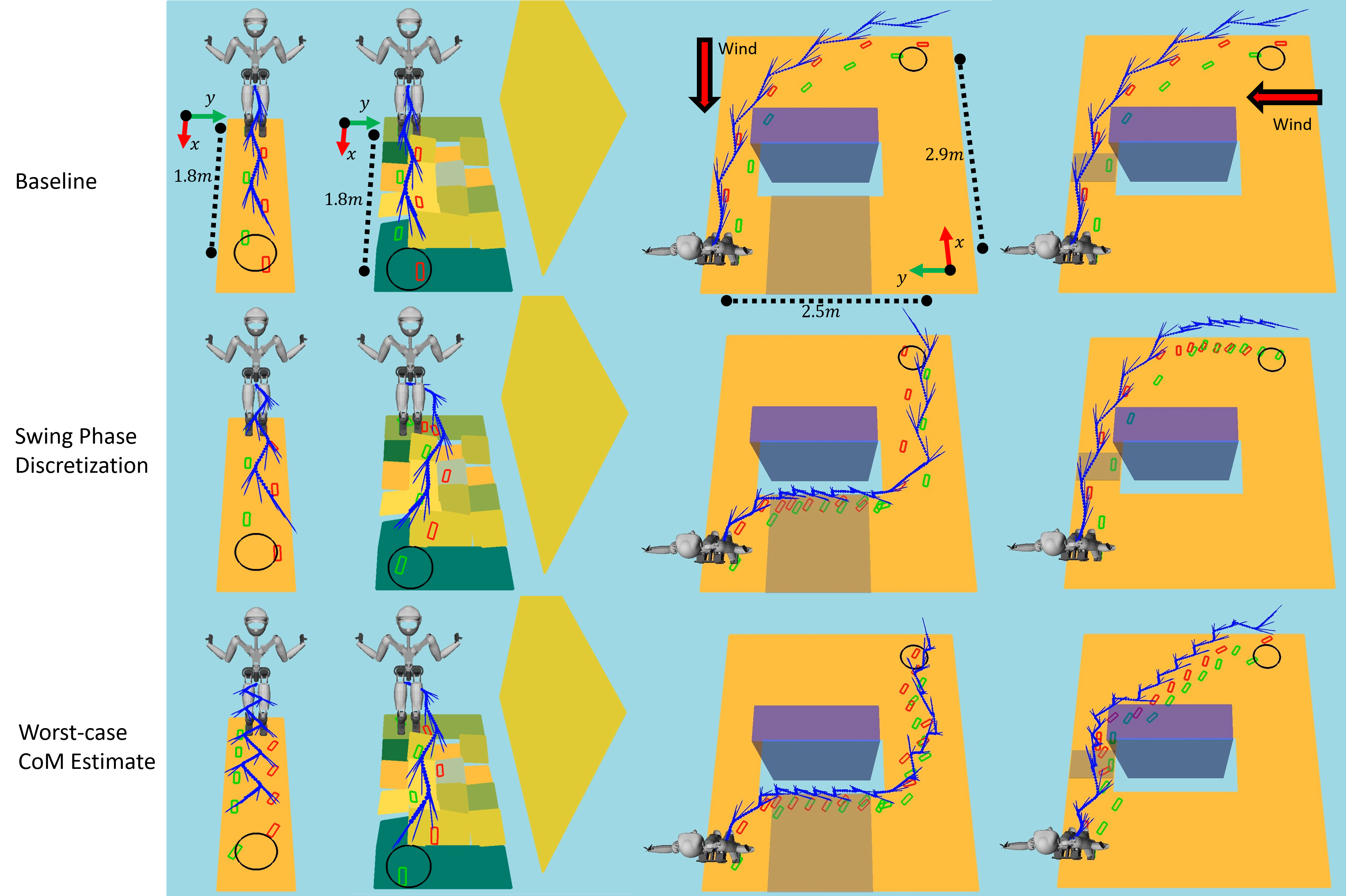}
	\vspace{-0.1in}
	\caption{From left to right: The planned footstep sequence in the narrow flat corridor, the rubble with wall, and the oil platform (wind in $-X$ and $+Y$ direcitons). The CoM trajectories returned by the kino-dynamic optimizer given the footstep sequences are shown in blue.}
	\vspace{-0.2in}
	\label{planning_result_big}
\end{figure*}

\section{Experiments}

We evaluate the performance of the proposed approaches in three test environments in simulation: a narrow, flat strip of ground, a field of rubble with an adjacent wall, and part of an oil platform, as shown in Figure \ref{planning_result_big}. For each test, we allow 1 minute planning time, and set $w_{s}=3$, $w_{cap}=1000$, $\gamma=0.1$ and the friction coefficient is $0.5$. We compare the proposed approaches with the baseline approach which only considers moving distance and step number ($w_{cap}=0$). For all test environments, we show the planned footstep sequences in Figure \ref{planning_result_big}, and summarize the quantitative results in Figure \ref{result_table}.

Since small disturbances can be handled by the robot's momentum controller, and do not require the planner to explicitly find capture motion to reject them, in the below experiment, we only consider the relatively rare but dangerous case that high disturbances $D_{\text{high}}$ act on the robot. Unless otherwise stated, we set the probability of those high disturbances happening within every time step (0.1 second) as $P(D_{\text{high}},0.1)=1\%$. To make the result easier to interpret, we let $P(\delta_{i},0.1), \ \delta_{i} \in D_{\text{high}}$ evenly divide $P(D_{\text{high}},0.1)$.

To evaluate the planned contact sequence, we first get its corresponding kinematic trajectory using the iterative kino-dynamic optimizer described in Section \ref{centroidal_momentum_opt}. Each trajectory is a discrete sequence of $\mathbf{q},\mathbf{\dot{q}}$ with time steps of 0.1 second. For each time step $t_{j}$ of the kinematic trajectory, including both swing and double support phases, we take the configuration as the initial kinematic state, and apply disturbances $\delta_{i} \in D_{\text{high}}(\mathbf{r}(t_{j}))$ one by one and check if the robot can capture itself using the approach described in Section \ref{evaluation_of_capturability}. For each disturbance $\mathbf{\delta_{i}}$, we first check if the condition is zero-step capturable, if not, we then check if it is one-step capturable with any of the capture poses generated using contact projection shown in Figure \ref{contact_projection}. In double support phase, when testing one-step capturability, we allow the robot to break one existing contact, and make contact at a capture pose. With the capturability of the robot for each time step - disturbance pair, we finally compute the probability that the robot finishes the path without falling due to external disturbance $P_{\text{success}}\left(T_{cp}\right)$ to evaluate the path quality.

We run the experiments on an Intel i7-8700K 3.7GHz CPU, and use an NVIDIA GeForce RTX 2080 GPU to speed up network queries for the Swing Phase Discretization approach. The neural networks are trained with Keras 2.2.4, and queried with Tensorflow 1.4 C++ API. The robot model we use is a Sarcos Humanoid robot.

\subsection{Prediction of Zero-Step and One-Step Capturability}

\begin{figure}[tp]
\medskip
\footnotesize
\begin{center}
\begin{tabular}{|c|c|c|c|}
\hline
Index & Precision & Recall & Accuracy \\ \hline
0     & 97.4\%    & 98.3\% & 97.8\%   \\ \hline
1     & 98.0\%    & 98.0\% & 98.0\%   \\ \hline
2     & 95.9\%    & 94.3\% & 95.2\%   \\ \hline
3     & 92.2\%    & 90.3\% & 91.3\%   \\ \hline
\end{tabular}
\end{center}
\vspace{-0.1in}
\caption{The neural networks' performance}
\label{nn_result_table}
\end{figure}

Figure \ref{nn_result_table} summarizes the performance of the neural networks in predicting capturability given an initial stance, a CoM position and a linear momentum. For each capture motion category, we train the network with $10^{5}$ examples, and test it with another $1000$ examples. Although all models perform well in predicting the capturability, the performance of predicting capture motions using palm contacts is worse than its counterpart using foot contact. This may be because capture motions using palm contacts are more likely to violate kinematic constraints and have higher variance in kinematic state, which cause them to be harder to learn.


\begin{figure*}[t]
\vspace{0.01in}
\footnotesize
\subfloat{
    \adjustbox{valign=t}{
    \hskip-0.14cm
\begin{tabular}{|c|c|c|c|c|c|c|c|}
\hline
\multirow{2}{*}{\begin{tabular}[c]{@{}c@{}}Test\\ Environment\end{tabular}} & \multirow{2}{*}{Approach} & \multicolumn{3}{c|}{\begin{tabular}[c]{@{}c@{}}Number of Failed\\ Time Step - Disturbance Pairs\end{tabular}} & \multirow{2}{*}{\begin{tabular}[c]{@{}c@{}}Step \\ Number\end{tabular}} & \multirow{2}{*}{\begin{tabular}[c]{@{}c@{}}Planning Time (s)\\ (First Solution/Best Solution \\ within the Time Limit)\end{tabular}} & \multirow{2}{*}{$P_{\text{success}}\left(T_{cp}\right)$} \\ \cline{3-5}
 &  & Total & Swing Phase & \begin{tabular}[c]{@{}c@{}}Double Support\\ Phase\end{tabular} &  &  &  \\ \hline
\multirow{3}{*}{\begin{tabular}[c]{@{}c@{}}\\Narrow Flat \\ Ground\end{tabular}} & Baseline & 36 & 14/40 & 22/60 & 5 & 0.52/0.52 & 83.49\% \\ \cline{2-8} 
 & \begin{tabular}[c]{@{}c@{}}Swing Phase\\ Discretization\end{tabular} & \textbf{15} & 4/40 & 11/60 & 5 & 0.88/2.13 & \textbf{92.75\%} \\ \cline{2-8} 
 & \begin{tabular}[c]{@{}c@{}}Worst-case \\ CoM Estimate\end{tabular} & 18 & 6/72 & 12/108 & 9 & 0.60/4.29 & 91.37\% \\ \hline
\multirow{3}{*}{\begin{tabular}[c]{@{}c@{}}\\Rubble with\\ Wall\end{tabular}} & Baseline & 89.80$\pm$2.79 & 35.40$\pm$2.06/80 & 54.40$\pm$3.72/120 & 5$\pm$0 & 0.54$\pm$0.01/0.54$\pm$0.01 & 79.83$\pm$0.56\% \\ \cline{2-8} 
 & \begin{tabular}[c]{@{}c@{}}Swing Phase \\ Discretization\end{tabular} & 17.80$\pm$7.30 & 16.60$\pm$5.46/128 & 1.20$\pm$2.39/192 & 8$\pm$0.63 & 1.17$\pm$0.84/13.74$\pm$4.58 & 95.84$\pm$1.65\% \\ \cline{2-8} 
 & \begin{tabular}[c]{@{}c@{}}Worst-case \\ CoM Estimate\end{tabular} & \textbf{11.6$\pm$1.36} & 11.6$\pm$1.36/112 & 0$\pm$0/168 & 7$\pm$0 & 0.58$\pm$0.07/1.09$\pm$0.16 & \textbf{97.14$\pm$0.33\%} \\ \hline
\multirow{3}{*}{\begin{tabular}[c]{@{}c@{}}\\Oil Platform\\ (Wind in $-X$\\ direction)\end{tabular}} & Baseline & 25 & 10/144 & 15/216 & 12 & 0.54/11.353 & 91.99\% \\ \cline{2-8} 
 & \begin{tabular}[c]{@{}c@{}}Swing Phase \\ Discretization\end{tabular} & \textbf{0} & 0/288 & 0/432 & 24 & 1.249/22.381 & \textbf{100\%} \\ \cline{2-8} 
 & \begin{tabular}[c]{@{}c@{}}Worst-case \\ CoM Estimate\end{tabular} & 1 & 1/384 & 0/576 & 32 & 0.582/30.728 & 99.67\% \\ \hline
\multirow{3}{*}{\begin{tabular}[c]{@{}c@{}}\\Oil Platform\\ (Wind in $+Y$\\ direction)\end{tabular}} & Baseline & 45 & 18/144 & 27/216 & 12 & 0.54/11.353 & 86.02\% \\ \cline{2-8} 
 & \begin{tabular}[c]{@{}c@{}}Swing Phase \\ Discretization\end{tabular} & 61 & 23/252 & 38/378 & 21 & 1.103/2.572 & 81.54\% \\ \cline{2-8} 
 & \begin{tabular}[c]{@{}c@{}}Worst-case \\ CoM Estimate\end{tabular} & \textbf{42} & 23/312 & 19/468 & 26 & 0.555/35.894 & \textbf{86.90\%} \\ \hline
\end{tabular}}}
\caption{The performance of each approach in all test environments. Note that there are 4 and 6 time steps in swing and double support phase, respectively. $P_{\text{success}}(T_{cp})$ is only affected by failed time step - disturbance pairs, so even some contact sequences are longer, its $P_{\text{success}}(T_{cp})$ can still be higher.}
\vspace{-0.2in}
\label{result_table}
\end{figure*}

\subsection{Narrow Flat Ground Test Environment}

In this test environment, we would like to show an intuitive result of how the robot can adjust its footstep placement to be more robust to external disturbances. We consider two lateral disturbances: $D_{\text{high}} = \left\{m[0,\pm0.6,0]^{T}\right\}\text{kg}\cdot\text{m/s}$. In this case, the most dangerous situation is when the robot shifts its CoM to one side, and the disturbance pushes in the same direction. In this situation, the robot mainly relies on zero-step capture motion to reject the disturbance. The proposed approaches make the robot increase the step width of the motion, which expands the support region in the $y$ direction, and hence makes the robot more stable.

\subsection{Rubble with Wall Test Environment}

In this test, the robot has to traverse through a rubble with a side wall, similar to the rubble environment used in the DARPA Robotics Challenge. We test for five randomly generated rubble surfaces with different tilt angles, and set $D_{\text{high}} = \left\{m[0,0.5,0]^{T}, m[0,0.6,0]^{T}, \right.$ $\left. m[0,0.7,0]^{T}, m[0,0.8,0]^{T}\right\}\text{kg}\cdot\text{m/s}$. Although the wall provides a wide space for the robot to capture itself using palm contacts, it is too far away for the robot to reach if the robot simply walks straight to the goal. The planner is able to incorporate this information, and adjust the path to be close to the wall, and achieves a much more robust footstep sequence under the disturbances.

\subsection{Oil Platform Test Environment}

This test demonstrates how the planner adapts to different sets of disturbances. We consider a part of an offshore oil platform with wind blowing. There are structures on the oil platform that can block the wind, but are not suitable for palm contacts, such as electronics and pipes. We first consider $D_{\text{high}} = \left\{m[-0.6,0,0]^{T}, m[-0.7,0,0]^{T}, \right.$ $\left. m[-0.8,0,0]^{T}\right\}\text{kg}\cdot\text{m/s}$, and the wind is blocked by the structure in the center, which creates a region without disturbance, shown in grey in Figure \ref{result_table}. We show that the proposed approaches leverage this region to produce low-risk contact sequences.

In another test, we consider a different wind direction with $D_{\text{high}} = \left\{m[0,0.6,0]^{T}, m[0,0.7,0]^{T}, m[0,0.8,0]^{T}\right\}\text{kg}\cdot\text{m/s}$. In this test, we show that the proposed approach is able to adapt to this change and produce a different contact sequence, shown in Figure \ref{planning_result_big}. However, this wind direction imposes great challenges to the planner because the robot will have to travel a long distance under the strong wind. This will create many high-cost edges, which drive predicted $P_{\text{success}}(T_{cp})$ low, and many paths look similarly costly in planning. Therefore, it is not easy for ANA* to reduce search space quickly. In this case, the Worst-case CoM Estimate approach and the baseline outperform the Swing Phase Discretization approach. The first reason is that the shortest path happens to be a good path in this case. The second reason is that Swing Phase Discretization approach branches each state much slower than the other approaches due to the large amount of network queries. Therefore, it failed to find a good solution within the time limit.

\subsection{Summary of the Planning Results}

In summary, we show that the proposed approaches generate contact sequences more robust to disturbance for the scenarios considered, except for the oil platform environment with $+Y$ wind direction where Worst-case CoM Estimate approach and the baseline have similar performance. Although Worst-case CoM Estimate simplifies the capturability check of each contact transition for higher efficiency, its performance is comparable to Swing Phase Discretization approach. In general, compared to the baseline, the proposed approaches take longer to plan a contact sequence. However, if we consider scenarios with shorter horizon, such as the narrow flat ground and rubble with wall environment, Worst-case CoM Estimate approach has planning time much shorter than the execution time, and could be used in a receding horizon fashion.

\section{Discussion}

\textcolor{black}{While the proposed approaches perform much better than the baseline, there still are time steps that the robot failed to reject the disturbances when following the footstep sequence generated with the proposed approaches. In addition to wrong predictions by the network, the disturbance rejection probability model could sometimes be misleading. In Section \ref{dusturbance_rejection_probability}, we define the disturbance rejection probability to depend on the number of feasible capture poses. Since many capture poses are similar, as shown in Figure \ref{contact_projection}, if there is a wrong prediction, the network is likely to have multiple wrong predictions given by similar capture poses. To improve the model, one possible direction for future work is to use ensemble learning to increase the prediction's robustness.}

In this work, we plan humanoid contact sequences which enable the robot to more easily capture itself under external disturbances. While the decision on where to place contacts is crucial for a successful capture, CoM position and centroidal momentum also play an important role. In our current approach, during planning, the CoM position and centroidal momentum of the robot in each state is determined by a neural network which learns from a dynamics optimizer without the information of the disturbances. The solution quality may increase if we includes CoM position and centroidal momentum as decision variables in the planner. However, this will significantly increase the branching factor, and slow down the planning.

While our approach is capable of finding footstep sequences that are more robust to potential
disturbances, it is still necessary to have a controller to react to disturbances
during execution and select the appropriate next contact in real-time. Several approaches
have been proposed to find the next contact location that helps stabilize a robot, such as in
\cite{Mason:2018vk}, but they often use a simplified model of the dynamics. It could be interesting
to extend the learning part of our approach to use it in a real-time controller in order
to remove the need for simplifying assumptions on the dynamics.

\section{Conclusion}
In this paper we addressed the problem of finding contact sequences that
are not only dynamically feasible but are also robust to external disturbances. It is the first time, to the best of our knowledge that a contact
planning algorithm explicitly considers the effect of external disturbances. 
In order to enable a fast evaluation of the capturability of a transition, we trained classifiers using neural networks, leading to a significant
speed-up in planning time. Experimental results demonstrate that our approach can
quickly find contact plans that are less susceptible to external disturbances, which leads to more robust behaviors when executed on a real robot.







\bibliography{references}{}
\bibliographystyle{unsrt}

\end{document}